\def\year{2020}\relax
\documentclass[letterpaper]{article} 
\usepackage{aaai20}  
\usepackage{times}  
\usepackage{helvet} 
\usepackage{courier}  
\usepackage[hyphens]{url}  
\usepackage{graphicx} 
\usepackage{graphicx}  
\usepackage{color}
\usepackage{comment}
\usepackage[utf8]{inputenc}
\usepackage{graphicx}
\usepackage{multirow}
\usepackage{verbatim}
\usepackage[table,xcdraw]{xcolor}

\title{Unclogging Our Arteries: \\
Using Human-Inspired Signals to Disambiguate Navigational Intentions}
\author{Justin Hart, Reuth Mirsky, Stone Tejeda, Bonny Mahajan, \\ \Large \textbf{Jamin Goo, Kathryn Baldauf, Sydney Owen, and Peter Stone}\textsuperscript{\rm 1} \\
\textsuperscript{\rm 1}
Department of Computer Science, \\
The University of Texas at Austin, Austin, TX, 78712 \\
hart@cs.utexas.edu, reuth@cs.utexas.edu, stonetejeda@utexas.edu, jgoo@utexas.edu, \\ bm.bonny@utexas.edu, kathrynbaldauf@utexas.edu, seowen@utexas.edu,pstone@cs.utexas.edu}

\begin{document}

\maketitle

\begin{abstract}
People are proficient at communicating their intentions in order to avoid conflicts when navigating in narrow, crowded environments. In many situations mobile robots lack both the ability to interpret human intentions and the ability to clearly communicate their own intentions to people sharing their space. This work addresses the second of these points, leveraging insights about how people implicitly communicate with each other through observations of behaviors such as gaze to provide mobile robots with better social navigation skills. In a preliminary human study, the importance of gaze as a signal used by people to interpret each-other's intentions during navigation of a shared space is observed. This study is followed by the development of a virtual agent head which is mounted to the top of the chassis of the BWIBot mobile robot platform. Contrasting the performance of the virtual agent head against an LED turn signal demonstrates that the naturalistic, implicit gaze cue is more easily interpreted than the LED turn signal.

\end{abstract}

\section{Introduction}
When robots and humans must navigate together in a shared space, conflicts may arise when they choose conflicting trajectories. Humans are able to disambiguate such conflicts between one another by communicating, often passively, through non-verbal communicative cues. The process by which people share spaces such as hallways involves cues such as body posture, motion trajectories, and gaze. In cases where this communication breaks down, the parties involved may do a ``Hallway Dance,''\footnote{\url{https://www.urbandictionary.com/define.php?term=Hallway}$\%$\url{20dance}} wherein they navigate into the same space - even doing so several times while trying to deconflict from each other's paths - rather than gracefully passing each other. This occurrence, however, is rare and socially-awkward for the participants.

Robots generate trajectories which can often be difficult for people to interpret, and generally communicate very little about their internal states passively \cite{Dragan-2013-7671}. This behavior can lead to situations similar to the hallway dance, wherein they clog the traffic arteries in confined spaces such as hallways or even in crowded, but open spaces such as atria. The Building-Wide Intelligence project \cite{khandelwal2017bwibots} at UT Austin intends to create an ever-present fleet of general-purpose mobile service robots. With multiple robots continually navigating our Computer Science Department, we have had many opportunities to witness these robots come into conflict with people when passing them in shared spaces. The most common type of conflict occurs when a human and a robot should simply pass each other in a hallway, but instead stop in front of each other; thus inconveniencing the human and possibly causing the robot to choose a different path.

In previous work \cite{fernandez2018passive}, our group sought to overcome this difficulty by incorporating LED turn signals onto the robot. It was found that the turn signals are not easily interpreted by participants, but that the introduction of a ``passive demonstration'' showing their use allows the signals to be understood. A passive demonstration is a training episode wherein the robot demonstrates the use of the turn signal in front of the user without explicitly telling the user that they are being instructed. In the case of our previous study, the robot simply makes a turn, using the turn signal, within the field of view of the user. Thus, the user has the opportunity to witness the signal before it is important for interaction with the robot, but is not explicitly told its purpose. However, limitations of this technique include that it demands that the robot recognize when it is first interacting with a new user, allowing it to perform the demonstration, and that an opportunity arises to perform such a demonstration before the signal must be used in practice.

This work designs and tests a more naturalistic signaling mechanism, hypothesizing that naturalistic signals will not require such a training period. Signaling mechanisms such as gaze or body language, mimicking human non-verbal communicative cues, may be far more easily understood by untrained users. Gaze is an important cue used to disambiguate human navigational intentions. A person will look in the direction that they intend to walk simply to assure that the path is safe and free of obstacles, but doing so enables others to observe their gaze. Observers can interpret the trajectory that the person performing the gaze is likely to follow, and coordinate their behavior. From this observation follows the design of a series of two studies. The first study is a human field experiment exploring the importance of gaze in the navigation of a shared space. The second study is a a human-robot study contrasting a robot using an LED turn signal with a gaze cue rendered on a virtual agent head. These studies support the hypotheses that gaze is an important social cue used when navigating shared spaces and that the interpretation of gaze as a naturalistic communicative cue is more clear to human observers than the artificial cue of a LED  signal when used in this context. A video demonstrating different test conditions and responses can be found online. \footnote{\url{https://youtu.be/mOwZo6uRREc}}

\section{Related Work}
Of recent interest to the robotics research community is the study of humans and robots navigating in a shared space \cite{Baraka2018,fernandez2018passive,murakami2002collision,ratsamee2013human,8525528,Szafir:2015:CDF:2696454.2696475,tamura2010smooth,watanabe2015communicating}. Our prior study introduced the concept of a ``passive demonstration,'' in order to disambiguate the intention of a robot's LED turn signal \cite{fernandez2018passive}. Baraka and Veloso (\citeyear{Baraka2018}) used an LED configuration on their CoBot to indicate a number of robot states - including turning - focusing on the design of LED animations to address legibility. They performed a study showing that the use of these signals increases participants' willingness to aid the robot. Szafir, Mutlu, and Fong (\citeyear{Szafir:2015:CDF:2696454.2696475}) equipped quad-rotor drones with LEDs mounted in a ring at the base, providing four different signal designs along this strip. They found that their LEDs improve participants' ability to quickly infer the intended motion of the drone. Shrestha, Onishi, Kobayashi, and Kamezaki (\citeyear{8525528}) performed a study similar to ours, in which a robot crosses a human's path in several different ways, indicating its motion intention with an arrow projected onto the floor using a color video projector. They found that in the scenario of a person and the robot passing each other in a hallway, similar to that presented in this paper, their method is effective in expressing the robot's intended motion trajectory. They intend to explore the use of shoulder-height turn signals in future work.

Gaze has been studied heavily in HRI \cite{admoni2017social}. It is a common hypothesis that gaze following is ``hard-wired'' in the brain \cite{emery2000eyes}. Generating gaze on the behalf of the robot is a naturalistic signal, emulating human behavior. A significant portion of human communication takes place implicitly, based on a combination of the context in which communication takes place and factors such as body language \cite{admoni2017social}. A robot navigating down a hallway, crossing a person's path and getting out of the way leveraging predictions based on gaze (such as those made in \cite{unhelkar2015human}) would be leveraging implicit social cues to coordinate its behavior. In this work, the robot generates a gaze signal to convey its intention to a person, despite the fact that doing so has no impact on its actual vision. This gaze signal can be contrasted with communicative signals designed specifically for non-humanoid robots and other devices \cite{cha2018survey}, such as LED turn signals. 

People rely heavily on non-verbal communication in social interactions \cite{argyle2013bodily,breazeal2005effects}. The present work builds on the demonstrated concept that humans infer other people’s movement trajectories from their gaze direction~\cite{nummenmaa2009ll}, and on the relationship between head pose and gaze direction \cite{kar2017review}. Norman (\citeyear{norman2009design}) speculated that bicycle riders know how to avoid collision with pedestrians since the latter group's members are consistent with their gaze.

Other works have dealt with visible change of posture when a pedestrian is about to change course, such as weight shifting, foot location and leg and pelvic rotations \cite{patla1999online,stirling2013examining}. Patla, Adkin, and Ballard's (\citeyear{Patla1999}) detailed description of whole-body kinematics during walking motion, was leveraged by Unhlelkar, Perez-D'Arino, Stirling, and Shah (\citeyear{unhelkar2015human}) to create a predictor for discretized human motion trajectories. Unhlelkar et al (\citeyear{unhelkar2015human}) found that head pose is a significant predictor of the direction that a person intends to walk. In their study, they discretized trajectories in terms of a decision problem of which target a person would walk towards.

Following a similar line of thought, Khambhaita, Rios-Martinez, and Alami (\citeyear{khambhaita2016head}) propose a motion planner which coordinates head motion to the path a robot will take $4$ seconds in the future. In a video survey in which their robot approaches a T-intersection in a hallway, they found that study participants were significantly more able to determine the intended path of the robot in terms of the left or right branch of the intersection when the robot used the gaze cue as opposed to when it did not. Using a different gaze cue, Lynch, Pettr{\'e}, Bruneau, Julien, Kulpa, Cr{\'e}tual, and Olivier (\citeyear{lynch2018effect}) performed a study in a virtual environment in which virtual agents established mutual gaze with participants during path-crossing events in a virtual hallway, finding no significant effect in helping participants to disambiguate their paths from those of the virtual agents. 

Of course, this gaze behavior extends beyond walking and bicycling. Recent work in our laboratory has studied the use of gaze as a cue for interacting with copilot systems in cars \cite{ROMAN18-Jiang,ICMI18-Jiang}, also with the aim of inferring the driver's intended trajectory. Gaze is also often fixated on objects being manipulated, which can be leveraged to improve algorithms which learn from human demonstrations \cite{saran2019understanding}. Though the use of instrumentation such as head-mounted gaze trackers or static gaze tracking cameras is limiting for mobile robots, recent work in the development of gaze trackers which work without such equipment \cite{saran2018human} may soon allow us to repeat the robot experiments presented in this paper with the robot reacting to human gaze, rather than only generating a gaze cue.

\section{Human Field Study}
This human ecological field study observes the effect of violating expected human gaze patterns while navigating a shared space. In this study, research confederates sometimes look opposite to the direction in which they intend to walk, violating the expectations observed by the work of Patla et al. (\citeyear{Patla1999}) and Unhlelkar et al. (\citeyear{unhelkar2015human}) by which head pose is predictive of trajectory. We hypothesize that causes problems in interpreting the navigational intent of the confederate, and can lead to confusion or near-collisions.

\subsection{Experimental Setup}

\begin{figure}[ht]
    \centering
    \includegraphics[width=4cm]{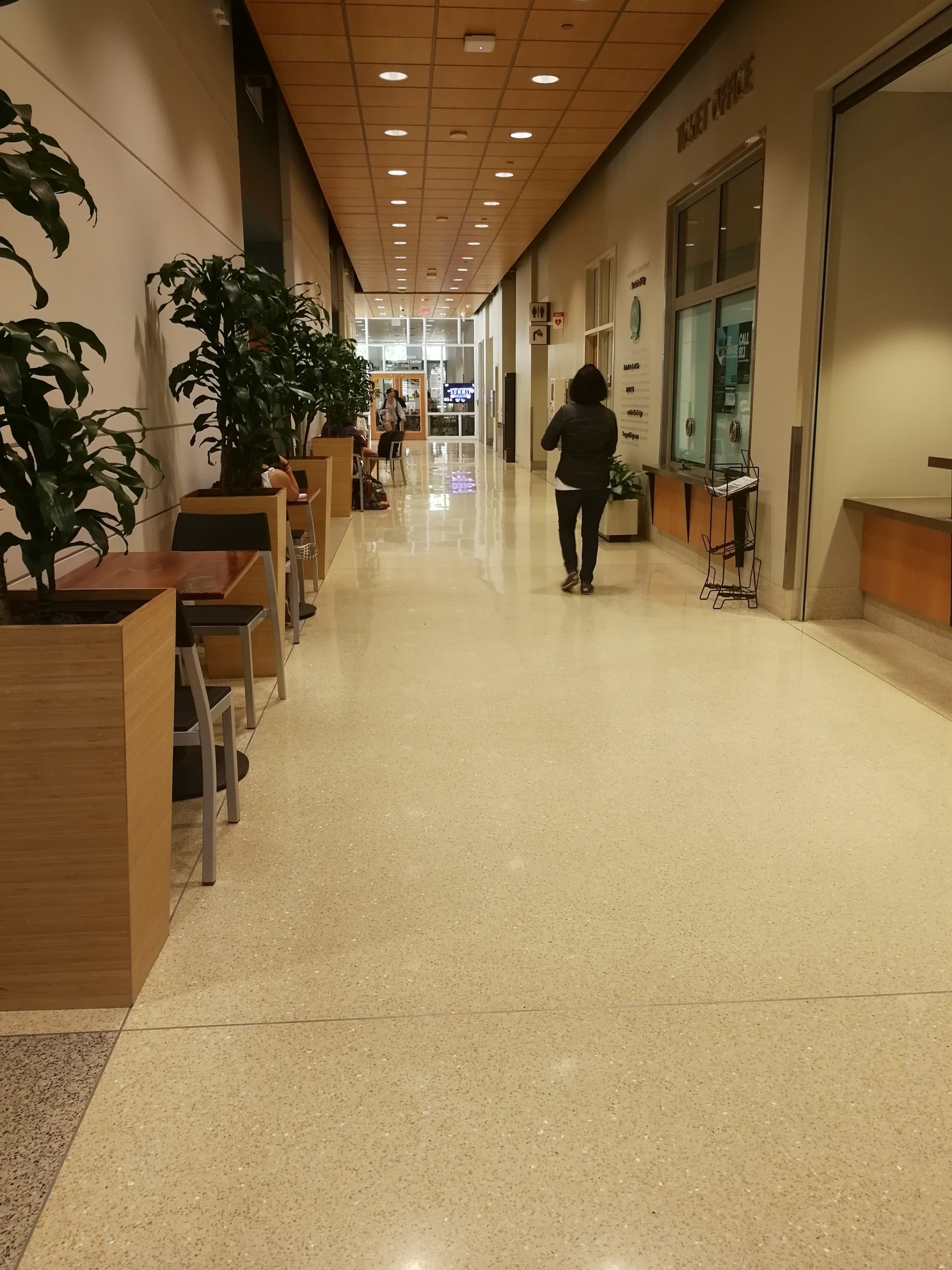}
    \includegraphics[width=5cm, angle=90]{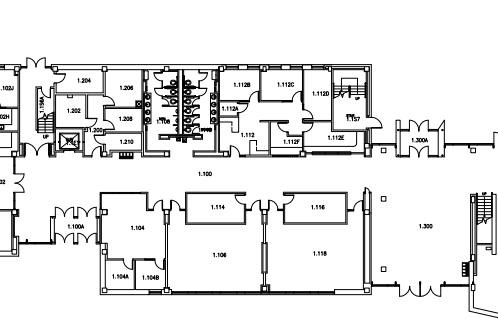}
    \caption{The hallway in which the human field experiment took place.}
    \label{fig:SAC}
\end{figure}

The Student Activity Center at UT Austin is a busy, public building hosting meeting rooms for student activities and a variety of restaurants in its food court. It has predictable busy times, centering around the schedule of class changes at the university. This study was performed in a busy hallway, Figure \ref{fig:SAC}, which becomes crowded during class changes. 

Prior to this study, two of the authors of this paper trained each other to proficiently look counter to the direction in which they walk, and acted as confederates who interacted with study participants. Both of the confederates who participated in this study are female. A third author acted as a passive observer to interactions between these confederates and other pedestrians walking through the hallway.

This experiment is organized as a $2\times2$ study, controlling whether the interaction occurs in a ``crowded'' or ``uncrowded'' hallway, and whether the confederate looks in the direction in which they intend to go (their gaze is ``congruent'') or opposite to this direction (their gaze is ``incongruent''). 
Here, ``crowded'' is defined as a state in which it is difficult for two people to pass each other in the hallway without coming within $2$m of each other. It can be observed that the busiest walkways at times form ``lanes'' in which pedestrians walk directly in lines when traversing these spaces. This study was not performed under these conditions, as walking directly toward another pedestrian would require additionally breaking these lanes, introducing another variable into the study.

The passive observer annotated all interactions in which the confederate and a pedestrian walked directly toward each other. If the confederate and the pedestrian collided with each other or nearly collided with each other, the interaction was annotated as a ``conflict.'' Conflicts are further divided into ``full'' collisions, in which the two parties bumped into each other; ``partial,'' in which the confederate and pedestrian brushed against each other; or ``shift,'' in which the two parties shifted to the left or right to deconflict each other's paths after coming into conflict.

\subsection{Results}
A total of $116$ interactions were observed ($65$ female / $51$ male), with $55$ in crowded conditions and $61$ in uncrowded conditions. The confederate looked in the incongruent direction in $38$ of the of the uncrowded interactions and $35$ of the crowded interactions. A one-way ANOVA found no significant main effect between the crowded and uncrowded conditions ($F=1.07655, p=0.301667$),  whether the confederate went to the pedestrian's right or left during the interaction ($F=2.35787, p=0.127423$), or based on gaze direction ($F=1.72727, p=0.191398$). Whether the gaze direction was congruent with the walking direction, however, was significant ($F = 4.11232, p=0.045$). A breakdown of conflicts based on the congruent condition versus the incongruent gaze condition can be found in Table \ref{tab:human}. These results support the hypothesis that humans use gaze to deconflict their navigational trajectories when crossing each other's paths in an ecologically-valid setting.

\begin{table}[ht]
\centering
\begin{tabular}{|c||c|c|}
\hline
\textbf{Conflict Type} & \textbf{Congruent} & \textbf{Incongruent}\\ \hline
Partial       & 8 (18\%)  & 9 (12\%)  \\ \hline
Quick Shift   & 5 (12\%)  & 24 (33\%) \\ \hline
Full          & 0 (0\%)        & 3 (4\%) \\ \hline \hline
Any Conflict & 13 (30\%)        & 36 (49\%) \\ \hline 
No Conflict   & 30 (70\%)  & 37 (51\%)  \\ \hline \hline
Total          & 43 (100\%) & 73 (100\%)  \\ \hline
\end{tabular}
\caption{Ratio of participants divided by congruence of gaze and walking direction contrasted with the type of conflict.}
\label{tab:human}
\end{table}

\section{Gaze as a Navigational Cue for HRI}
\label{sec:gaze_cue}
Motivated by the use of gaze as a naturalistic cue for implicitly communicating navigational intent, we engineered a system in which the BWIBot uses a virtual agent head to communicate the direction it intends to steer toward in order to deconflict its trajectory from that of a human navigating a shared space by looking in the direction that it intends to navigate toward. In these experiments, a human starts at one end of a hallway, and the robot starts at the other end. The human is instructed to traverse the hallway to the other end, and the robot also autonomously traverses it. As a proxy for measuring understanding of the cue, the number of times that the human and robot come into conflict with each other is measured for two conditions: one in which the robot uses a turn signal to indicate the side of the hallway that it intends to pass the person on,\footnote{The robot's motion is intended to be similar to a car changing lanes.} and one in which it uses a gaze cue to make this indication. This experiment is motivated by prior work \cite{fernandez2018passive} which introduced the concept of a ``passive demonstration'' in order to overcome the challenge of people not understanding the intention of the turn signal. The question asked here is, will the gaze cue, based on natural behavior and implicit communication, outperform the artificial signal of the LED turn signal.

\begin{figure}[ht]
    \vspace*{2mm}
    \centering
    \includegraphics[width=8cm]{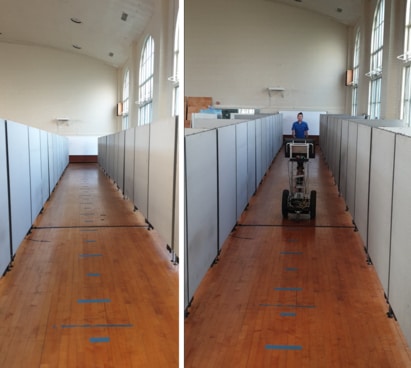}
    \caption{The hallway constructed for this experiment (left) and a demonstration of the experiment during execution (right).}
    \label{fig:hallway}
\end{figure}

\begin{figure}[ht]
    \centering
    \includegraphics[width=8cm]{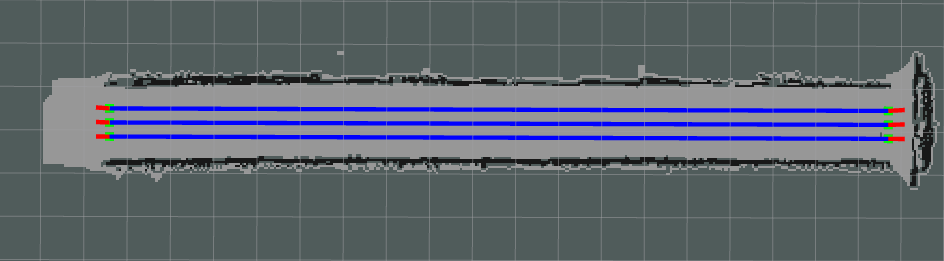}
    \caption{The robot's mental map of the hallway.}
    \label{fig:lanes}
\end{figure}

This study is set in a test hallway built from cubicle furniture, shown in Figure \ref{fig:hallway}. The hallway is $17.5$m long by $1.85$m wide. The system models the problem of traversing the hallway as one in which the hallway is divided into three lanes, similar to traffic lanes on a roadway, which is illustrated in Figure \ref{fig:lanes}. If the human and the robot are within $1$m of each other when they cross each other's path, they are considered to be in conflict with each other. This distance is based on the $1$m safety buffer engineered into the robot's navigational software, which also causes the robot to stop. If the human and the robot both start in the middle lane, at opposing ends of the hallway, with the navigational goal of traversing to the opposing ends of the hallway, the signal is intended to indicate an intended lane change on behalf of the robot, so that the human may interpret that signal, and shift into the opposing lane if necessary. 

\subsection{Gaze Signal}

\begin{figure}
    \vspace*{2mm}
    \centering
    \includegraphics[width=4cm]{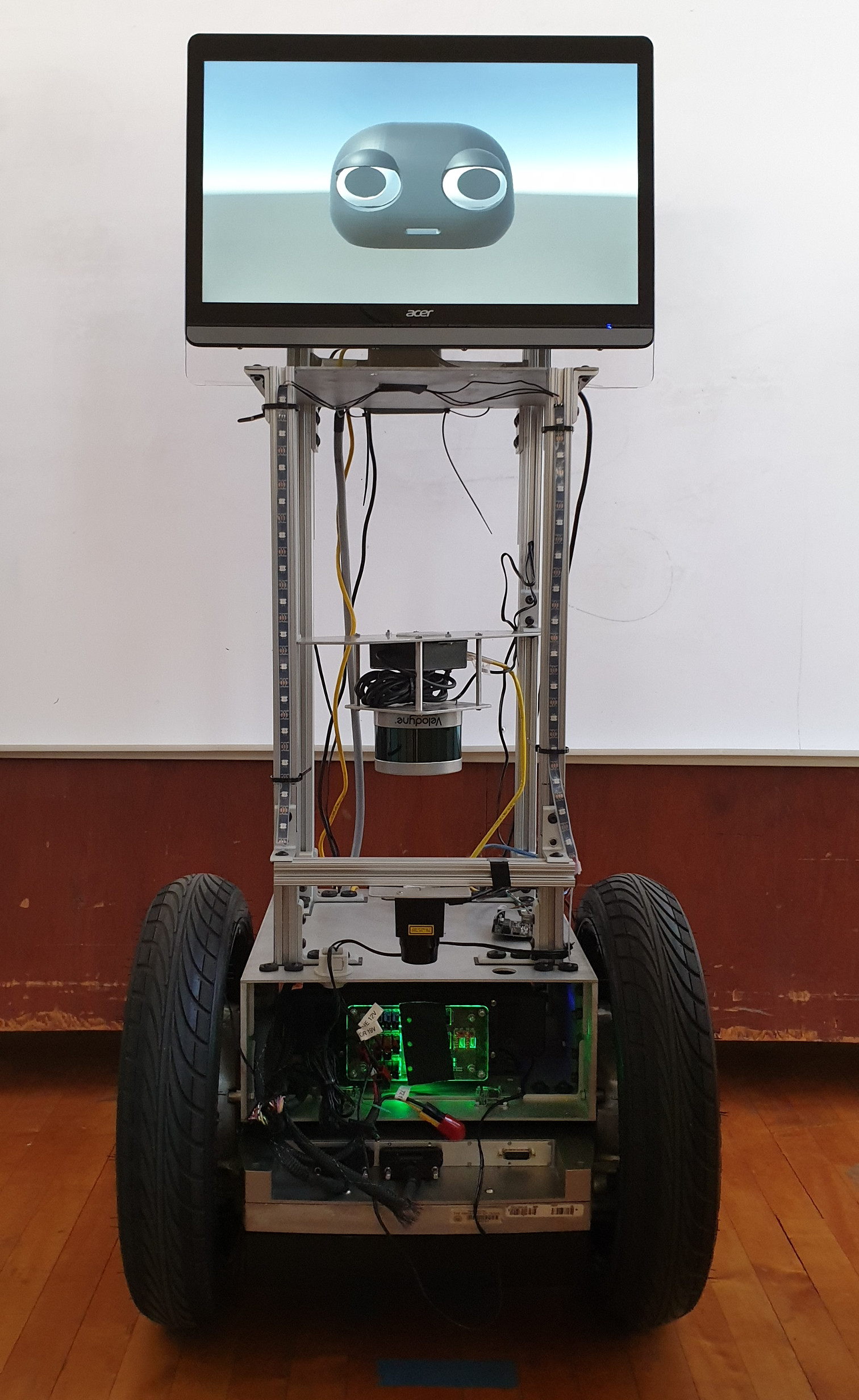}
    \includegraphics[width=4cm]{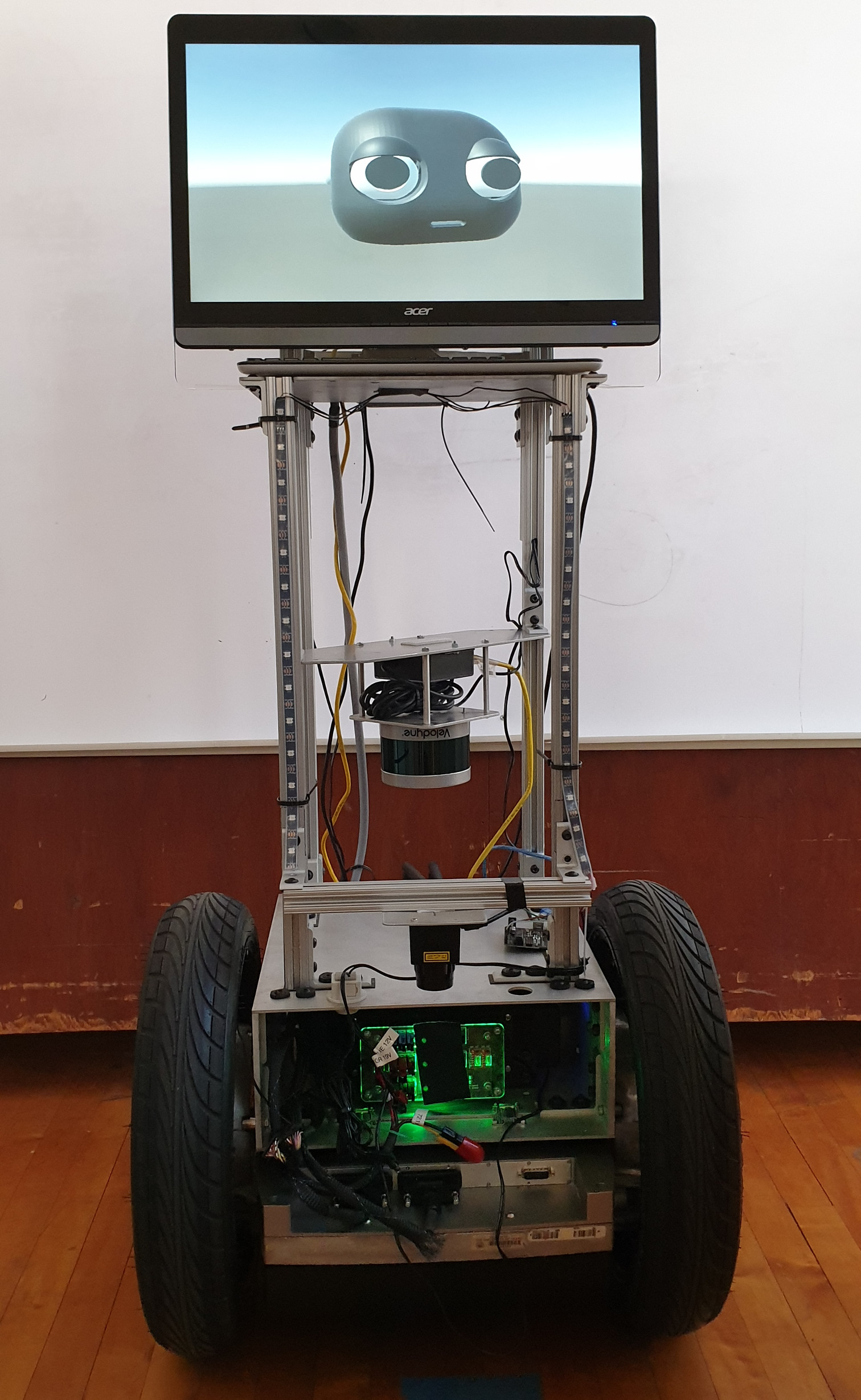}
    \caption{Gaze Signal. Left: Neutral; Right: Signaling to the left lane.}
    \label{fig:conditions_gaze_1}
\end{figure}

To display gaze cues on the BWIBot we developed a 3D-rendered version of the Maki 3D-printable robot head.\footnote{\url{https://www.hello-robo.com/maki}} The decision to use this head is motivated by the ability to both render it as a virtual agent and, in future work, to 3D print and assemble a head. The virtual version of the head was developed by converting the 3D-printable STL files into Wavefront .obj files and importing them into the Unity game engine.\footnote{\url{https://unity.com/}} To control the head and its gestures, custom software was developed using ROSBridgeLib.\footnote{\url{https://github.com/MathiasCiarlo/ROSBridgeLib}} The head is displayed on a $21.5$ inch monitor mounted to the front of the robot. When signaling, the robot turns its head $16.5^{\circ}$ and remains in this pose. The eyes are not animated to move independently of the head. The head turn takes $1.5$ seconds. These timings and angles were hand-tuned and pilot tested on members of the laboratory. The gaze signal can be seen in Figure \ref{fig:conditions_gaze_1}.

\subsection{LED Signal}
\begin{figure}[ht]
    \centering
    \includegraphics[width=4cm]{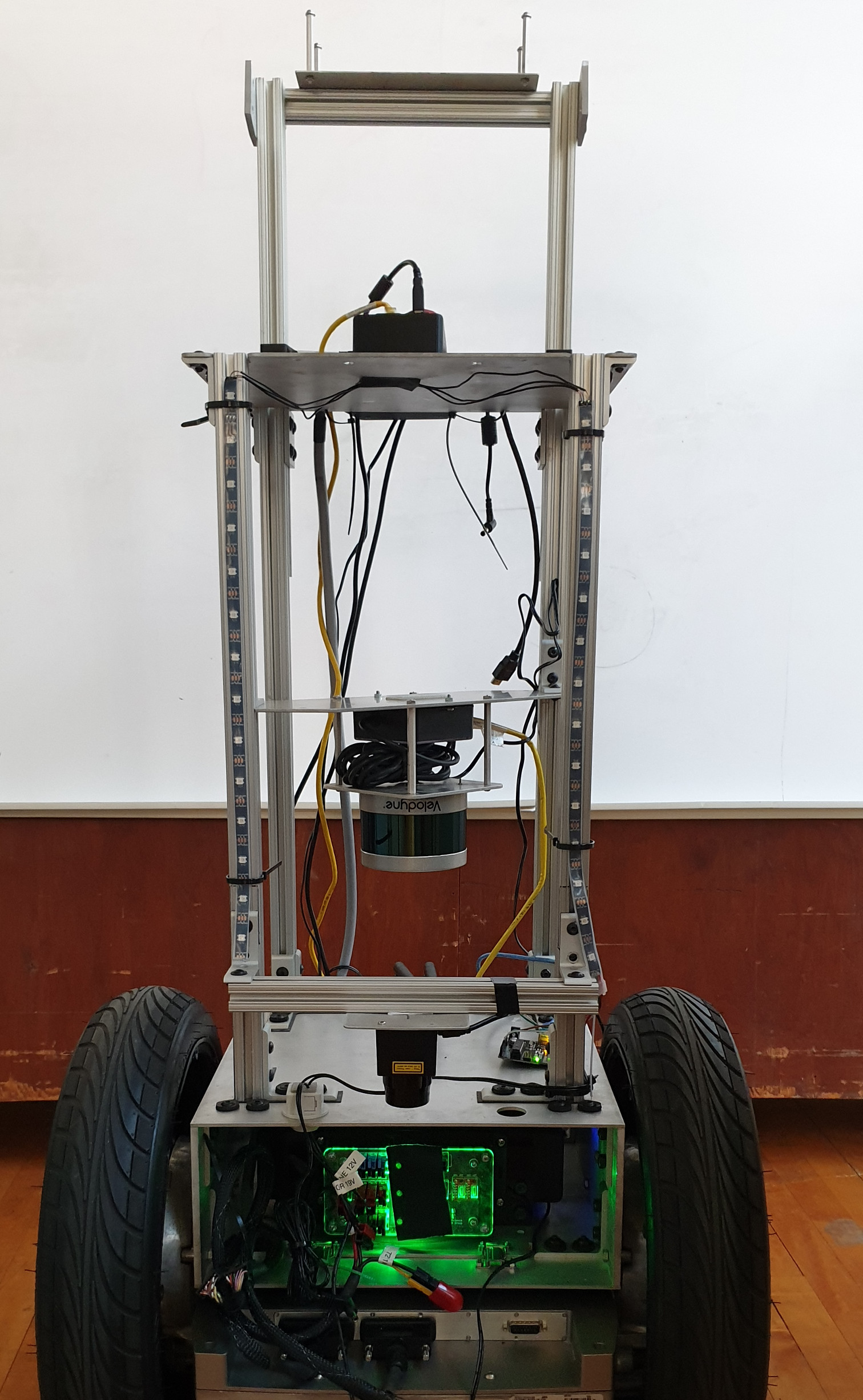}
    \includegraphics[width=4cm]{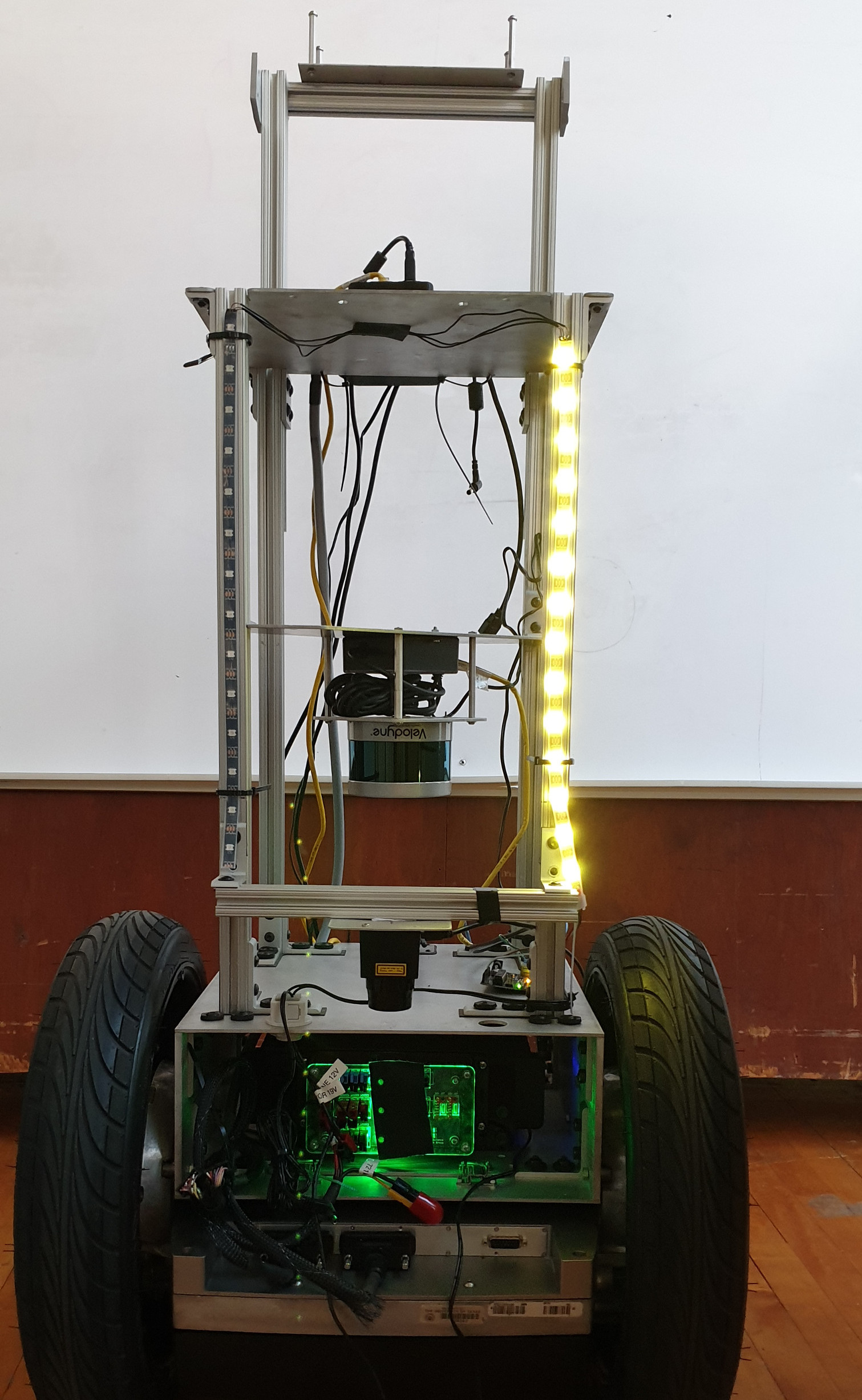}
    \caption{LED signals. Left: Neutral; Right: Blinking to signal a move to the left lane.}
    \label{fig:conditions_LED}
\end{figure}

The LED cue is a re-implementation of the LED turn signals from \cite{fernandez2018passive}. Strips of LEDs $0.475$m long with $14$ LEDs line the 8020 extrusion on the chassis of the front of the BWIBot. They are controlled using an Arduino Uno microcontroller, and blink twice per second with $0.25$ seconds on and $0.25$ seconds off each time they blink. In the condition that the LEDs are used, the monitor is removed from the robot. The LED signals can be seen in Figure \ref{fig:conditions_LED}.

\subsection{Experimental Setup}
\label{sec:exp}

To test the effectiveness of gaze in coordinating navigation through a shared space, we conducted a human-robot interaction study in the hallway test environment. After obtaining informed consent and, optionally, media release, participants are guided to one end of the hallway, where the robot is already set up at the opposite end. The participant is instructed to navigate to the opposing end of the hallway, and both the participant and the robot start in the ``middle lane,'' as per the model in Section \ref{sec:gaze_cue}. When the participant starts walking down the hallway, the robot is also started. The study presented here uses an inter-participant design, in which each participant sees exactly one of the two cues -- gaze or LED. Each participant traverses the hallway with the robot exactly once in order to avoid training effects. After completing the task of walking down the hallway, each participant responds to a brief post-interaction survey.

To assure that the study results are reflective of the robot's motion signaling behavior, rather than the participants' motion out of the robot's path, the study is tuned to give the participant enough time to get out of the robot's way by reacting to its gaze or LED cue. The problem is modeled using three distances: $d_{signal}$, $d_{execute}$ and $d_{conflict}$. The distance $d_{signal}$ ($4$m) is the distance at which the robot signals its intention to change lanes, which is based on the distance at which the robot can accurately detect a person in the hallway using a leg detector \cite{leigh2015person} and its on-board LiDAR sensor. The distance at which the robot will execute its turn, $d_{execute}$ ($2.75$m), was hand-tuned to be at a range at which it is unlikely that the participant will have time to react to the robot's gaze or LED cue. This design is so that if the participant has not already started changing lanes by the time the robot begins its turn, it is highly likely that the person and robot will experience a conflict. Thus, this study tests interpretation of the queue, not reaction to the turn. The distance at which the robot determines that its motion is in conflict with that of the study participant is $d_{conflict}$, which is set to $1$m. This design is based on the safety buffer used when the robot is autonomously operating in our building.

In addition to using the three ranges to control the robot's behavior, the robot also always moves itself into the ``left'' lane. This choice was made because, in North America, pedestrians usually walk to the right of each other in order to deconflict each other's paths. Preliminary testing of the combination of these two behaviors showed that the pedestrians and the robot came into conflict $100\%$ of the time. As such, when participants move out of the robot's path, it can be attributed mainly to the robot's signaling.

The post-interaction survey comprises $44$ questions, consisting of $8$-point Likert and cognitive-differences scales, and one free-response question. Five demographic questions on the survey ask whether people in the country where the participant grew up drive or walk on the left or right-hand side of the road and about their familiarity with robots. Ten questions concern factors such as the clarity of the signaling method used by the robot. Finally, twenty nine questions ask about perceptions of the robot in terms of personality factors such as selfishness or whether the robot is perceived as threatening; factors of safety and usefulness; and other factors such as appropriate environments for the robot, such as the workplace or home. The free response prompts the following to participants in the LED condition, ``There would be a better position for the signals, and it is:'' The full survey as given to the participants is available online 
\footnote{\url{https://docs.google.com/forms/d/1aTVx_cdhLMZPosKktS5FxqJZbanXiXJs6Ey8XQh7tQg/prefill}}.

\subsection{Results}
We recruited $38$ participants ($26$ male / $12$ female), ranging in age from $18$ to $33$ years. The data from $11$ participants is excluded from our analysis. Last-minute software changes before the first day of testing led to a software failure that was only detected after $8$ participants had completed the experiment (after a participant asked, ``What face?'' we investigated). Two of the participants, early on, were robotics students who had read the previous paper \cite{fernandez2018passive} and were familiar with the hypotheses of the study. After reviewing participant recruitment and carefully repairing the software and re-piloting the study, testing resumed. The final excluded data comes from a participant who failed to participate in the experimental protocol.

The remaining pool of participants includes $11$ participants in the LED condition and $16$ in the gaze condition. Table \ref{tab:robot} shows the results from the robot signaling experiment in these two conditions. A pre-test for homogeneity of variances confirms the validity of a one-way ANOVA for analysis of the collected data. A one-way ANOVA shows a significant main effect ($F=7.711$, $p=0.002$). Post-hoc tests     of between-groups differences using the Bonferroni criteria show significant mean differences between the gaze group and the LED group (gaze versus LED: $md=-0.50, se=0.14, p=0.004$), but no significant mean difference between the gaze and LED conditions ($md=-0.08, se=0.15, p=1.00$). None of the post-interaction survey responses revealed significant results. A video of two example interactions can be found at\\
\noindent \url{https://youtu.be/MHT3NU3NueM}. These results support the hypothesis that the robot's gaze can be more readily interpreted in order to deconflict its trajectory from that of a person navigating in a shared space.

\begin{table}[ht]
\centering
\begin{tabular}{|c||c|c|c|}
\hline
             \textbf{Conflict Type}  & \textbf{LED}         & \textbf{Gaze}           \\ \hline
Conflict       & 11         & 8                  \\ \hline
No Conflict    & 0           & 8                   \\ \hline \hline
Total & 11 & 16    \\ \hline
\end{tabular}
\caption{Number of participants divided by condition and whether there was a conflict in the person's and robot's trajectories.}
\label{tab:robot}
\end{table}

\section{Discussion \& Future Work}
\label{sec:discussion}
The goal of these studies is to evaluate whether a naturalistic, implicit communicative gaze cue outperforms a more synthetic LED turn signal in coordinating the behavior of people and robots when navigating a shared space. This work follows previous work, finding that LED turn signals are not readily interpreted by people when interacting with the BWIBot, but that a brief, passive demonstration of the signal is sufficient to disambiguate its meaning \cite{fernandez2018passive}. This study asks whether gaze can be used without such a demonstration.

The human ecological field study presented here  validates the use of gaze as an implicit communicative cue for coordinating trajectories. Gaze may even be a more salient cue than a person's actual trajectory in this interaction.

In the human-robot study that follows, we compared the performance of an LED turn signal against a gaze cue presented on a custom virtual agent head. In this condition, the robot turns its head and ``looks'' in the direction of the lane that it intends to take when passing the study participant. Our results demonstrate that the gaze cue significantly outperforms the LED signal in preventing the human and robot from choosing conflicting trajectories. We interpret this result to mean that people naturally understand this cue when the robot makes it, transferring their knowledge of interactions with other people onto the template of their interaction with the robot.

The gaze cue does not perform perfectly in the context of this study. There are several potential contributing factors. The first is that, while the entire head rotates, the eyes do not move to focus on any point in front of the robot. There are also minor errors in the construction of the 3D model that make it look slightly unnatural at times.\footnote{These have been addressed and will be rectified in a future study.} The distance window between signaling and performing the lane-change is also briefer in this study than in the previous study \cite{fernandez2018passive}. This change in timing is because we had to change robots\footnote{The robot is a similar, custom, BWIBot, with a slightly different base and sensor suite.} due to electrical problems. The leg detector \cite{leigh2015person} does not work as reliably on the updated platform. In the previous study, participants are signaled at $7$m, as opposed to $4$m in this study, as a result of issues with leg detection. Finally, interpreting gaze direction on a virtual agent head may be difficult due to the so-called ``Mona Lisa Effect'' \cite{ruhland2015review}. In follow-up studies, we intend to both tune the behavior of the head, and contrast its performance against a 3D printed version of the same head. The decision to use a virtual agent version of the Maki head is driven by our ability to contrast results in future experiments (upon construction of the hardware) between virtual agents and robotic heads.

Subtle differences between the robot's behavior and human behavior may make the signal's intention ambiguous, or it may simply be that robot gaze is interpreted differently from human gaze. Additionally, the lack of physical embodiment of the head possibly plays heavily into the performance of displayed cues. A detailed survey of the gaze literature discussing these factors is beyond the scope of this paper, but many are addressed by Admoni and Scassellati (\citeyear{admoni2017social}). Significant future work to analyze these factors is in the planning phases.

The overall results of this study are highly encouraging. Many current-generation service robots avoid what may be perceived by their designers as overly-humanoid, perhaps unnecessary facial features and expressions. However, the findings in this work indicate that such features may be more readily interpreted by people interacting with these devices, and thus be highly beneficial.



\section*{Acknowledgements}
This work has taken place in the Learning Agents Research Group (LARG) at the Artificial Intelligence Laboratory, The University of Texas at Austin.  LARG research is supported in part by grants from the National Science Foundation (IIS-1637736, CPS-1739964, IIS-1724157), the Office of Naval Research (N00014-18-2243), Future of Life Institute (RFP2-000), Army Research Laboratory, DARPA, and Lockheed Martin.  Peter Stone serves on the Board of Directors of Cogitai, Inc.  The terms of this arrangement have been reviewed and approved by the University of Texas at Austin in accordance with its policy on objectivity in research.


\bibliographystyle{aaai}
\bibliography{lbib}

\end{document}